# An Evolutionary-Based Approach to Learning Multiple Decision Models from Underrepresented Data

Vitaly Schetinin, Dayou Li, Carsten Maple
Computing and Information Systems Department, University of Bedfordshire, Luton, UK, LU1 3JU
{vitaly.schetinin, dayou.li, carsten.maple}@beds.ac.uk

**Abstract**

*Abstract – The use of multiple Decision Models (DMs) enables to enhance the accuracy in decisions and at the same time allows users to evaluate the confidence in decision making. In this paper we explore the ability of multiple DMs to learn from a small amount of verified data. This becomes important when data samples are difficult to collect and verify. We propose an evolutionary-based approach to solving this problem. The proposed technique is examined on a few clinical problems presented by a small amount of data.*

## 1. Introduction

The use of multiple Decision Models (DMs) enables to enhance the generalization ability of the models. At the same time the use of multiple DMs allows users to evaluate the confidence in decision making. These properties are important for such applications as medical diagnostics when data collected for learning diagnostic models are represented by a small amount of patients' cases which have been reliably verified. Ideally, the DMs should satisfy the following requirements: they should provide the maximal performance; they should provide the estimates of confidence in decisions, and finally the decision models should be interpretable for domain experts [1, 2].

Such models can be learnt from a set of verified data which should be large enough to represent a problem. However, in practice domain data collected by users are often underrepresented to be used for learning decision models with an acceptable predictive accuracy. For example, in clinical practice the domain data can be collected from a small number of patients because of the difficulty of their diagnostic verification. In such cases the resultant models become unacceptably dependent on the variations in the collected data. Moreover, the use of cross-validation techniques aiming to mitigate the overfitting problem becomes inefficient on a small amount of data, affecting the ability of models to generalize. However, the generalization ability can be still evaluated within leave-one-out technique [2 - 4].

Fortunately, the loss of generalization ability can be compensated by averaging over multiple models' outcomes without the common cross-validation technique [3 - 5]. The use of multiple models also allows the confidence in decisions to be evaluated in terms of the consistency of the models' outcomes [3].

In this paper we anticipate that the generalization ability of DMs can be further enhanced by using the leave-one-out technique while the DMs learn from training data. We hope that this technique will allow us to mitigate the negative effect of overcomplicating DMs, which usually affects their ability to generalize [6].

The use of evolutionary-based algorithms for learning DMs of the growing complexity was shown efficient on a small amount of data [4 - 6]. To allow DMs to be gradually grown, these algorithms employ two basic operations: generating candidate-models and selecting the best of them as described in [4 - 10].

Within our approach DMs can be considered as multiple semantic networks which are logically interpretable [1, 2]. The models consist of processing units linked with attributes or outputs of the previous units. Each model learns from the domain data independently from each others in order to enhance the models' diversity. Making an ensemble of models diverse we can therefore improve its generalization ability by averaging over the multiple models. The use of multiple models allows us also to naturally estimate the confidence in decisions.

In many practical cases, domain data can be represented by many attributes while only a small amount of these data is available to collect. Learning DMs from such data cannot be efficient, and therefore the DMs loose the ability to generalize. However, the efficiency of the learning from a small amount of data can be improved by decomposing models with many attributes into several reference models with a few attributes [11 - 14]. The evolutionary-based method proposed in [15] has proved the efficiency of such approach on several problems.

Thus, the novelty of our approach to improve DMs learnt from a small amount of data is that we use the leave-one-out technique for selection of the reference models composing a DM. To make DMs interpretable, the reference models are represented in a logical form. Within our approach evolutionary composition of the reference models allows us to combine them into DM ensemble providing a better performance.

The remainder of this paper is organized as follows. Section 2 describes our evolutionary-based approach to learning DMs from domain data. Section 3 presents the cases of applications of our method to the clinical problems, and finally section 4 concludes the paper.

## 2. An Evolutionary Technique for Learning DMs from Underrepresented Data

In general, when models learn from a small amount of data, their performance becomes poor due to the loss of generalization ability [1 - 4]. The performance becomes especially poor when models are dependent on many arguments. However, in such cases the performance of models can be improved by decomposing their functions into several functions dependent on a smaller number of arguments, particularly, on two arguments [14 - 15]. In this paper our attention is focused on logical functions because of their interpretability.

Let $g_k(u_1, u_2)$, $k = 1, \ldots, 10$, denote all 10 logical functions of two arguments $u_1$ and $u_2$. Then according to the theorem about decomposition, any function of $m$ arguments $x_1, \ldots, x_m$, $f(x_1, \ldots, x_m)$ can be written as follows:

$$f(x_1,...,x_m) = g(g(...,g(x_i,x_j)),g(...)), \quad (1)$$

where $x_i, x_j$, $i \neq j$, $j = 1, \ldots, m$, are arguments of logical function $f$.

The logical functions of two arguments are listed in Table 1, where operators +, *, and ~ represent logical OR, AND, and NOT, respectively.

Table 1. Logical functions of two arguments.

| # | Functions |
|---|---|
| 1 | $g = u_1 + u_2$ |
| 2 | $g = \sim u_1 + u_2$ |
| 3 | $g = \sim(u_1 + u_2)$ |
| 4 | $g = u_1 + \sim u_2$ |
| 5 | $g = u_1 * u_2$ |
| 6 | $g = \sim u_1 * u_2$ |
| 7 | $g = \sim(u_1 * u_2)$ |
| 8 | $g = u_1 * \sim u_2$ |
| 9 | $g = \sim u_1 * u_2 + u_1 * \sim u_2$ |
| 10 | $g = \sim u_1 * \sim u_2 + u_1 * u_2$ |

Attributes or arguments $x_1, \ldots, x_m$ can be nominal or numeric. The nominal variables can be easily convertible into logical, while the numeric attributes need to be converted with minimal losses of information.

The simplest way of converting numeric attributes into logical is the use of a threshold technique [3]. However such conversion can loose important information.

The losses can be reduced if an attribute can be taken in combination with other attributes which determine a so-called context of the problem [1 - 5]. However, the determination of such context even for a few attributes is not a trivial task because of the combinatorial problem.

### 2.1. Training Models

Within our approach model units are trained one-by-one and then added to the model while the model performance increases. Therefore the number of processing units can be associated with the complexity of a model which increases gradually. For training we need to find such threshold $Q$ and model M, being represented as decomposition (1), which provide the best performance for a given complexity of model M.

To convert numeric attributes we can attempt to search for a desired threshold $Q$ in the context of the current DM by exploiting more than one attributes. The search procedure aims to minimise entropy of a model unit as follows. For attribute $A$ and set of training data, $S$, the value of entropy $H$ is:

$$H(S) = -\sum_{i=1}^{r} P_i \log_2(P_i), \quad (2)$$

where $r$ is a given number of classes, and $P_i$ is the probability of occurrence of class $C_i$.

The search procedure creates a candidate model M each input of which is linked to either attribute or the previous model. In this case the probability $P$ is calculated as a ratio $|S_i|/|S|$, where $S_i$ is the portion of training samples assigned by model M to class $C_i$, and $|S|$ is the total number of training samples. For the given threshold $Q$ and model (logical function of two arguments) $M_k$, the set $S$ is divided into two subsets $S_1$ and $S_2$ for which we can calculate conditional entropy $H(A, S | Q, M_k)$:

$$H(A,S|Q,M_k) = \frac{|S_1|}{|S|}H(S_1) + \frac{|S_2|}{|S|}H(S_2). \quad (3)$$

For a reasonably large number of the random samples from a set of attribute's values, this technique can find a threshold $Q^*$ and a logical function $M^*$ for which conditional entropy (3) is near minimal:

$$(Q^*, M^*) = \underset{Q_j \in S, 1 \leq k \leq 10}{\arg\min} (H(A, S | Q_j, M_k)). \quad (4)$$

During the search, the candidate values of $Q_j$ are drawn from a uniform distribution $U$ for a given number of samples, $l$, as follows:

$$Q_j \sim U(S), j = 1,..., l. \quad (5)$$

Thus, for a reasonably large number $l$, the above search procedure can find the solution $(Q^*, M^*)$.

### 2.2. Selection of Models

The DMs trained on a small amount of data can be selected by the number of errors on the training data. However, such selection favours overfitted DMs with a poor ability to generalise. To enhance the generalisation ability, we can attempt to select the DMs by using the leave-one-out cross validation technique allowing us to evaluate the generalisation ability in terms of average number of errors. For a small amount of data, this technique runs for a reasonable computational time.

Selecting the DMs, we need also exclude those DMs, which have different model's structures but provide the same outcomes; such DMs are redundant and useless repetitions, so-called tautologies. Within our approach both selection goals are achieved as follows.

Let $\mu$, $\mu_1$, and $\mu_2$ be the average numbers of errors admitted by a candidate model, M, and the previous models $M_1$ and $M_2$, respectively. The inputs of model M are linked with the outputs of models $M_1$ and $M_2$. Then the candidate model is selected if the following condition is met:

$$\mu < \min(\mu_1, \mu_2). \qquad (6)$$

Indeed, if a candidate model M admits fewer errors than models $M_1$ and $M_2$, then the generalisation ability of model M becomes better. Applying this heuristic rule iteratively to each candidate model, we therefore can achieve a near-optimal generalisation ability of a DM. In theory, the maximum of this ability cannot be guaranteed by using heuristics search.

The above rule can also effectively prevent the acceptance of the tautological DMs. Indeed, if model M is tautological, then it repeats one of models $M_1$ or $M_2$, and the above rule will reject such model.

### 2.3. Stopping Criteria

The learning process continues while the performance of DMs is improved. When the performance stabilises, we can assume that the trained DMs provide the best generalisation ability. Further performance improvement becomes unlikely and almost all candidate-models will be rejected by the selection criterion (6).

In general, the evolution process terminates due to one of the following reasons:

1) If an acceptable level of errors was achieved.
2) If the given number of unsuccessful attempts of improving the performance was exceeded.
3) If the given complexity of DMs was exceeded.

According to these criteria, one or more DMs in the last generation can provide a minimal number of errors. When there are several DMs, it seems reasonable to select from them those which have a minimal complexity, i.e., the DMs which comprise a minimal number of processing units.

Such selection allows us to discard overcomplicated DMs and use the remaining models $M_1$, …, $M_m$ of a minimal complexity, where $m$ is the number of models. Each DM was learnt from the training data independently from each other, and therefore we can arrange these DMs in an ensemble in order to enhance the reliability of decisions; e.g., the final decision can be made by the majority voting. Ensemble of DMs allows us also to evaluate the confidence in decisions.

### 2.4. Confidence in Decisions

When the resultant ensemble consists of several DMs, their outcomes calculated for a given input **x** can be inconsistent. For real-world problems, such inconsistency may be caused by noise or corruption in data. Therefore, we can determine the confidence in decisions in terms of the consistency $\chi$ calculated as a ratio $m_i/m$, where $m_i$ is the number of models voted for class $i$. Clearly, the estimates of confidence range between 0.5 and 1 (0.5 is the least confident, and 1.0 is the most confident).

## 3. Application to Clinical Problems

In this section first we describe our experiments with learning the DMs to differentiate Infectious Endocarditis (IE) from System Red Lupus (SRL) on the data collected in the Penza Hospital Rheumatology Department, see [15] for details. Second, we describe our experiments with the UK Trauma Data [16] and two data sets from the UCI ML Repository [17], the SPECT Heart and Wisconsin Prognostic Breast Cancer (WPBC).

To differentiate the IE and SRL, an expert collected 18 verified cases represented by the results of 24 clinical and laboratory tests which are commonly used for distinguishing these diseases. Among these 24 tests, 7 are represented by numeric values and the remaining 17 by nominal values. In total, the data set consists of 36 cases.

The Trauma data are represented by 16 variables: 11 categorical and 5 numerical. The SPECT data comprise 22 variables, all categorical. The WPBC data are represented by 33 variables, all numerical. To deal with underrepresented data in our experiments we use a small amount of data for training, namely, 50 data samples with an equal ratio of negative and positive outcomes.

We compare the proposed Evolving Decision Model (EDM) technique with a common Artificial Neural Network (ANN) ensemble technique. Each ANN is trained by back-propagation. The comparisons are made in terms of the performances provided by best single models (BS) and ensembles (E) within two-fold cross validation. The ensembles comprise 5000 ANNs, each including 10 hidden neurons, trained within leave-one-out validation technique to reduce the negative effect of overfitting to underrepresented data. To make the ANN ensemble effective, each ANN is randomly initialised and exploits a random set of variables.

For comparison we enable the proposed EDM technique to collect up to 5000 models as well. The complexity of models is made restricted to achieve the best performance; the complexity is simply counted by the number of model's units. In our experiments the complexity was restricted to 7, 10, and 20 units for the mentioned data sets.

Fig. 1 shows the fitness function values (the solid line in the upper plot) and complexity of models (the solid line in the lower plot) over the number of units accepted during the evolutionary learning. The dashed line in the upper plot depicts the performance of units within leave one-out technique, and the dashed line in the lower plot depicts the maximal complexity of models. From this figure we can see that the fitness function reaches a maximum very fast and then becomes to be slightly oscillated around 80%. The complexity of models is also grown fast and reaches a maximum around 6 units.

Fig. 2 depicts the performances of single models (in grey) and the performance of the ensemble (in black) over the number of models accepted during the learning. It is important that from this figure we can observe that the performance of an ensemble tends to increase.

The results of our experiments are listed in Table 2 presenting the mean and standard deviation values of the performances calculated over 5 runs within 2-fold cross validation. From this table we can see that the proposed EDM technique is superior to the ANN ensemble technique in terms of the predictive accuracy.

The additional advantage of the proposed EDM technique is that we can derive the best models from an ensemble to create a diagnostic table which clinicians can conveniently use to make decisions and evaluate their consistency. As an example, for the above IE and SRL differentiation problem we found the nine DMs, one of which is represented below in a form of if-then rules:

**if** *leukocytes* ($x_2$) are less than 6.2 **and**
*circulating immune complex* ($x_5$) is less than 130 **and**
*articular syndrome* ($x_8$) is absent **and**
*anhelation* ($x_{11}$) is absent **and**
*erythema* ($x_{13}$) is absent **and**
*noises in heart* ($x_{14}$) are absent **and**
*hepatomegaly* ($x_{15}$) is absent **and**
*myocarditis* ($x_{16}$) is absent,
**then** the diagnose is the IE.

Figure 3 shows the ranks of the eight attributes selected to be used in the DMs. The most important contribution is made by attributes $x_8$ and $x_{14}$, whilst attributes $x_5$ and $x_{15}$ make the weakest contribution. Therefore we can attempt to exclude the remaining attributes from the DMs keeping their performance high.

For a given input, the outcomes of the DMs may be inconsistent, and the final decision is made by the majority voting. The resultant diagnostic table has been applied to over 200 patients, and the misdiagnosed rate was less than 2%.

Our experiments were run in Matlab 7 with Intel Core 2 Duo CPU E675 2.66 GHz. The creation of a typical ensemble of DMs usually took about 10 min of the computations.

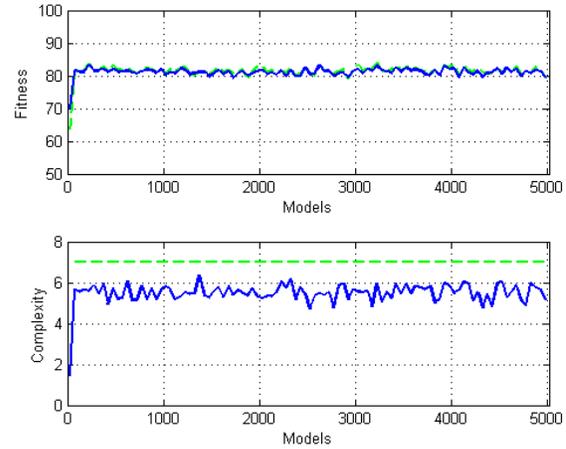

Fig. 1. Fitness function value (upper plot) and complexity (lower plot) over the number of units.

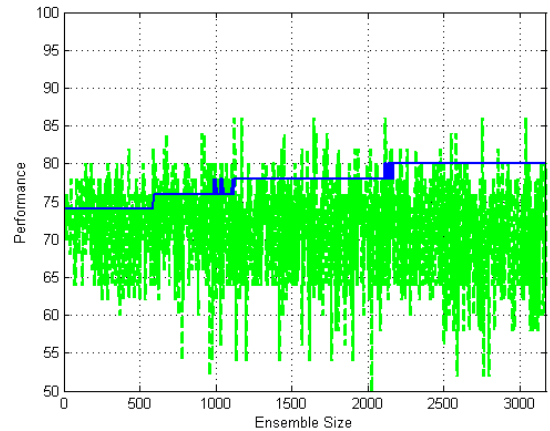

Fig. 2. Performances of single models (in gray) and an ensemble (in black) over the number of models.

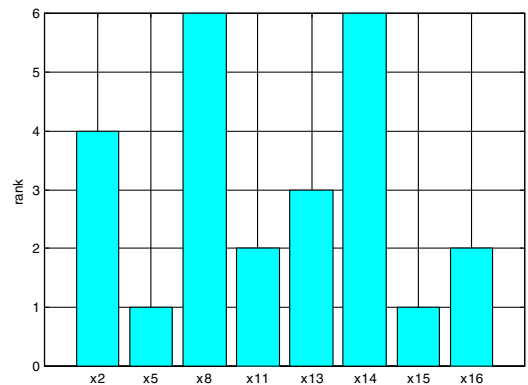

Fig. 3. Ranks of the attributes.

Table 2. Performances of the ANN and EDM techniques.

| Data | Fold | ANN | | EDM | |
|---|---|---|---|---|---|
| | | BS | E | BS | E |
| Trauma | 1 | 70.8±1.1 | 78.8±1.1 | 77.6±3.8 | **81.6±2.2** |
| | 2 | 77.2±1.8 | 80.8±1.1 | 75.6±3.8 | **83.2±3.0** |
| SPECT | 1 | 64.0±3.7 | 64.4±2.0 | 73.2±4.2 | **77.6±5.9** |
| | 2 | 62.0±3,2 | 64.0±1.4 | 69.6±2.2 | **72.8±1.8** |
| WPBC | 1 | 72.5±1.1 | 66.8±1.8 | 63.7±7.2 | **70.4±7.0** |
| | 2 | 64.2±2.3 | 66.3±0.9 | 64.2±4.0 | **67.1±9.6** |

## 4. Conclusions and Discussion

We have presented a new evolutionary-based approach to learning DMs from underrepresented data which frequently appear in practice. We aimed to find the DMs proving the best performance while keeping a form interpretable for users.

Based on an evolutionary approach, the proposed technique starts to learn the DMs consisting of one processing unit and then step-by-step evolves the DMs while their performance increases. Each processing unit has a pair of inputs that allows the uncertainty of model parameters learnt from the data to be reduced.

In theory, the proposed technique can provide a near-optimal complexity of models. However in practice, when dealing with underrepresented data, we cannot use a substantial portion of data for validation and, as a consequence, we needed to restrict the maximum of models' complexity. This restriction can be viewed as an additional technique for mitigating the overfitting problem.

Within our approach we combined the leave-one-out validation technique with the complexity restriction to achieve a near-optimal performance of ensemble. Our experiments on clinical datasets show that the proposed technique outperforms a common ANN ensemble technique in term of predictive accuracy. Additionally, the proposed technique is shown capable of excluding redundant clinical tests from decision models while their performance is kept high.

## 5. References

[1] J. Sowa, *Knowledge Representation: Logical, Philosophical, and Computational Foundations*. Brooks & Cole Publishing, Pacific Grove, CA, 2000.

[2] J. Doyle, "A Truth Maintenance System", *Artificial Intelligence*, 12, 1979.

[3] W. Kloesgen, and J. Zytkow (eds.), *Handbook of Data Mining and Knowledge Discovery*, Oxford University Press, New York, NY, USA, 2002.

[4] H. Madala, and A. G. Ivakhnenko, *Inductive Learning Algorithms for Complex Systems Modeling*, CRC Press Inc.: Boca Raton, 1994.

[5] J. A. Müller, and F. Lemke, *Self-Organizing Data Mining. Extracting Knowledge from Data*, Trafford Publishing, Canada, 2003.

[6] V. Schetinin, "Pattern Recognition with Neural Network", *Optoelectronics, Instrumentation and Data Processing*, Allerton Press, 2, 75-80, 2000.

[7] T. Back, D. Fogel, and Z. Michalewicz (eds), *Handbook of Evolutionary Computation*, IOP Publishing, Oxford University Press, 1997.

[8] A. Freitas, *Data Mining and Knowledge Discovery with Evolutionary Algorithms*, Springer-Verlag, 2002.

[9] W. Kwedlo, and M. Kretowski, "Discovery of Decision Rules from Databases: an Evolutionary Approach", *Principles of Data Mining and Knowledge Discovery, Second European Symposium (PKDD'98)*, Nantes, France, September 23-26, 1998. Springer Lecture Notes in Computer Science 1510, 1998.

[10] N. Nikolaev and H. Iba, "Automated Discovery of Polynomials by Inductive Genetic Programming", J. Zutkow, and J. Ranch (eds.) *Principles of Data Mining and Knowledge Discovery (PKDD'99)*, Springer, Berlin, 1999.

[11] G. Cybenko, "Approximation by Superpositions of a Single Function", *Mathematics of Control, Signals and Systems,* 2, 1989.

[12] R. Hecht-Nielsen, "Kolmogorov Mapping Neural Network Existence theorem", *Proc. IEEE First International Conference on Neural Networks*, San Diego, 1987.

[13] J. Schmidhuber, "Discovering Neural Nets with Low Kolmogorov Complexity and High Generation Capability", *Neural Networks*, 10, 1997.

[14] L. Kurkova, "Kolmogorov's Theorem and Multilayer Neural Networks", *Neural Networks*, 5, 1992.

[15] V. Schetinin, and A. Brazhnikov, "Extracting Decision Rules Using Neural Networks", *Biomedical Engineering*, Kluwer Academic, 1, 2000.

[16] Trauma Audit and Research Network. Available online: http://www.tarn.ac.uk

[17] The Machine Learning Repository of the University of California, Irvin. Available online: http://archive.ics.uci.edu/ml/datasets.html